%% file: icassp.tex
\documentclass{article}
\usepackage{spconf,amsmath,graphicx,subfig, balance}

\title{Learning Perception and Planning with Deep Active Inference}
\name{Ozan \c{C}atal\thanks{Ozan \c{C}atal is funded by a Ph.D. grant of the Flanders Research Foundation (FWO). This work is supported by AI Research Flanders} \qquad Tim Verbelen \qquad Johannes Nauta \qquad Cedric De Boom \qquad Bart Dhoedt}
\address{IDLab\\
Department of Information Technology at Ghent University - imec}
\input{math_commands.tex}

\begin{document}
%
\maketitle
\begin{abstract}
Active inference is a process theory of the brain that states that all living organisms infer actions in order to minimize their (expected) free energy. However, current experiments are limited to predefined, often discrete, state spaces. In this paper we use recent advances in deep learning to learn the state space and approximate the necessary probability distributions to engage in active inference. 
\end{abstract}
\section{Introduction}
\label{sec:intro}

Active inference postulates that action selection in biological systems, in particular the human brain, is actually an inference problem where agents are attracted to a preferred prior state distribution in a hidden state space~\cite{Friston2017}. To do so, each living organism builds an internal generative model of the world, by minimizing the so-called free energy. The idea of active inference stems from neuroscience~\cite{Friston2010,Friston2006} and has already been adopted to solve different control and learning tasks~\cite{Friston2012a, Friston2013, Friston2017}. These experiments however typically make use of manually engineered state transition models and predefined, often discrete, state spaces.

In this paper we show that we can also learn the state space and state transition model, by using deep neural networks as probability density estimators. By sampling from the learnt state transition model, we can plan ahead minimizing the expected free energy, trading off goal-directed behavior and uncertainty-resolving behavior. 

\section{Active Inference}
\label{sec:ai}

Active inference states that every organism or agent entertains an internal model of the world, and implicitly tries to minimize the difference between what it believes about the world and what it perceives, hence minimizing its own variational free energy~\cite{Friston2010}. Moreover, the agent believes that it will minimize its expected free energy in the future, in effect turning action selection into an inference problem. This boils down to optimizing for two distinct objectives. On the one hand the agent actively samples the world to update its internal model of the world and better explain observations. On the other hand the agent is driven to visit preferred states that it believes a priori it will visit---a kind of global prior--- which carry little expected free energy.

Formally, we assume an agent entertains a generative model $P(\tilde{\vo}, \tilde{\vs}, \tilde{\va}, \pi)$ of the environment, which specifies the joint probability of observations, actions and their hidden causes, where actions are determined by some policy $\pi$. If the environment is modelled as a Markov Decision Process (MDP) this generative model factorizes as:
\begin{equation}
   \label{eq:factor}
   \small
   \begin{split}
   P(\tilde{\vo}, \tilde{\vs}, \tilde{\va}, \pi) &=  P(\pi)P(\vs_0)
   \\
   &\quad \prod_{t=1}^{T}P(\vo_t|\vs_t)P(\vs_{t}|\vs_{t-1}, \va_{t-1})P(\va_{t-1}|\pi)
   \end{split}
\end{equation}

\noindent The free energy is then defined as: 
\begin{equation}
   \label{eq:free-energy}
   \begin{split}
      F &= \E_Q [ \log Q(\tilde{\vs}) - \log P(\tilde{\vo}, \tilde{\vs}, \tilde{\va})] \\
        &= \KL (Q(\tilde{\vs}) \Vert P(\tilde{\vs}, \tilde{\va} | \tilde{\vo}) ) - \log P(\tilde{\vo}) \\
        &= \KL (Q(\tilde{\vs}) \Vert P(\tilde{\vs}, \tilde{\va})) - \E_{Q} [ \log P(\tilde{\vo} | \tilde{\vs})]
   \end{split}
\end{equation}

\noindent where $Q(\tilde{\vs})$ is an approximate posterior distribution. The second equality shows that the free energy is minimized when the KL divergence term becomes zero, meaning that the approximate posterior becomes the true posterior, in which case the free energy becomes the negative log evidence. This can also be rewritten as the third equality, which is the negative evidence lower bound (ELBO) that also appears in the variational autoencoders (VAE) framework \cite{Kingma13, Rezende14}.

In active inference, agents infer the actions that will result in visiting states of low expected free energy. They do this by sampling actions from a prior belief about policies according to how much expected free energy that policy will induce. Formally, this means that the probability of picking a policy is given by \cite{Schwartenbeck2018}:
\begin{equation}
   \label{eq:G}
   \begin{split}
      P(\pi) &= \sigma(-\gamma G (\pi)) \\
      G(\pi) &= \sum_{\tau} G(\pi, \tau)
   \end{split}
\end{equation}
where $\sigma$ is the softmax function with precision parameter $\gamma$, which governs the agent’s goal-directedness and randomness in its behavior. Here $G$ is the expected free energy at future timestep $\tau$ when following policy $\pi$~\cite{Schwartenbeck2018}:
\begin{align}
   \label{eq:G-complete}
   \begin{split}
      G(\pi, \tau) &= \E_{Q(\vo_{\tau}, \vs_{\tau} \vert \pi)} [ \log Q(\vs_{\tau} \vert \pi) - \log P(\vo_{\tau}, \vs_{\tau} \vert \pi)] \\
                   &= \E_{Q(\vo_{\tau}, \vs_{\tau} \vert \pi)} [ \log Q(\vs_{\tau} \vert \pi) - \log P(\vo_\tau \vert \vs_\tau, \pi) \\ 
                   & \quad - \log P(\vs_\tau \vert \pi)] \\
                   &= \KL (Q(\vs_\tau | \pi) \Vert P(\vs_\tau)) + \E_{Q(\vs_\tau)} [ H(\vo_\tau \vert \vs_\tau)]
  \end{split}
\end{align}
We used $Q(\vo_\tau,\vs_\tau \vert \pi) = P(\vo_\tau \vert \vs_\tau)Q(\vs_\tau \vert \pi)$ and that the prior probability $P(\vs_\tau \vert \pi)$ is given by a preferred state distribution $P(\vs_\tau)$. This results into two terms: a KL divergence term between the predicted states and the prior preferred states, and an entropy term reflecting the expected ambiguity under predicted states. Action selection in active inference thus entails:

\begin{enumerate}
   \item Evaluate $G(\pi)$ for each policy $\pi$
   \item Calculate the belief $P(\pi)$ over policies
   \item Infer the next action using $P(\pi)P(\va_{t+1} \vert \pi)$
\end{enumerate}

\section{Deep Active Inference}
\label{sec:deepai}

For the agent's model, we use deep neural networks to parameterize the various factors of equation~(\ref{eq:factor}): i.e. the transition model $p_{\theta}(\vs_t | \vs_{t-1}, \va_{t})$ and the likelihood distribution $p_{\xi}(\vo_{t}|\vs_{t})$. Also the approximate posterior is parameterized by a neural network $p_{\phi}(\vs_t | \vs_{t-1}, \va_{t}, \vo_t)$. All distributions are parameterized as i.i.d multivariate Gaussian distributions, i.e.~the outputs of the neural networks are the means $\mu$ and standard deviations $\sigma$ of each Gaussian. Sampling is done using the reparameterization trick, computing $\mu + \epsilon \sigma$ with $\epsilon \sim N(0,1)$, which allows for backpropagation of the gradients. Minimizing the free energy then boils down to minimizing the following loss function:
\begin{align}
   \label{eq:model-obj}
   \begin{split}
   \forall t &: \underset{\phi, \theta, \xi}{\text{minimize}}: -\log p_{\xi}(\vo_{t}|\vs_{t}) \\
   & + \KL (p_{\phi}(\vs_t | \vs_{t-1}, \va_{t-1}, \vo_t) \Vert p_{\theta}(\vs_t | \vs_{t-1}, \va_{t-1}))
   \end{split}
\end{align}

\begin{figure}[t!]
    \centering
    \includegraphics[width=2.8in]{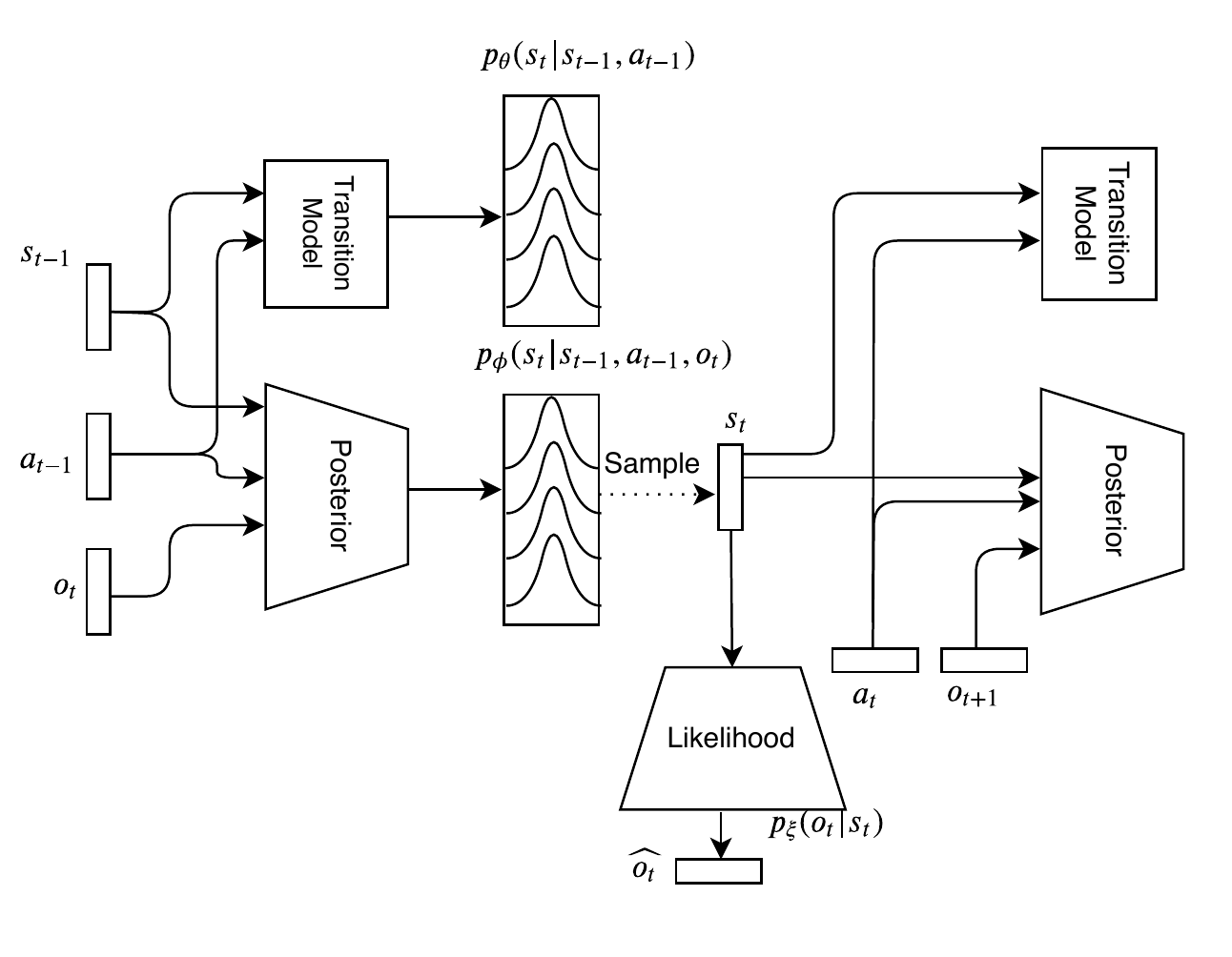}
    \caption{We train simultaneously a transition model $p_{\theta}(\vs_t | \vs_{t-1}, \va_{t})$, an approximate posterior distribution model $p_{\phi}(\vs_t | \vs_{t-1}, \va_{t}, \vo_t)$, and a likelihood distribution $p_{\xi}(\vo_{t}|\vs_{t})$ model, by minimizing the variational free energy.}
    \label{fig:approach}
\end{figure}

Figure~\ref{fig:approach} shows an overview of the information flow between the transition model, approximate posterior and likelihood neural networks. To engage in active inference using these models, we need to estimate $G(\pi,\tau)$, which involves estimating $Q(\vs_\tau | \pi)$. As our model takes a state sample as input, and only estimates the state distribution of the next timestep, the only way to get an estimate of the state distribution at a future timestep $\tau > t + 1$ is by Monte Carlo sampling. Concretely, to infer $P(\pi)$, we sample for each policy $N$ trajectories following $\pi$ using $p_{\theta}$. This results in $N$ state samples $\hat{\vs}_{\tau}$, for which we can get $N$ observation estimates $\hat{\vo}_{\tau}$ via $p_{\xi}$. To be able to calculate the KL divergence and entropy, we use a Gaussian distribution fitted on the samples' mean and variance. We then estimate the expected free energy for each policy from current timestep $t$ onward as follows:

{
\small
\begin{align}
   \label{eq:Gest}
   \hat{G_t}(\pi) &= \!\sum_{\tau=t+1}^{t+K}\! \KL (\mathcal{N}(\mu_{\hat{\vs}_{\tau}},\sigma_{\hat{\vs}_{\tau}}) \Vert P(\vs_\tau)) + \frac{1}{\rho} H(\mathcal{N}(\mu_{\hat{\vo}_{\tau}}, \sigma_{\hat{\vo}_{\tau}})) \nonumber \\
        & \;\; + \sum_{\pi^\prime} \sigma(-\gamma \hat{G}_{t+K}(\pi^\prime)) \hat{G}_{t+K}(\pi^\prime)
\end{align}
}

The first summation term looks $K$ timesteps ahead, calculating the KL divergence between expected and preferred states and the entropy on the expected observations. We also introduce an additional hyperparameter $\rho$, which allows for a trade-off between reaching preferred states on the one hand and resolving uncertainty on the other hand. The second summation term implies that after $K$ timesteps, we continue to select policies according to their expected free energy, hence recursively re-evaluating the expected free energy of each policy at timestep $t+K$. In practice, we unroll this $D$ times, resulting into a search tree with an effective planning horizon of $T = K \times D$. 

\section{Experiments}
\label{sec:experiments}

\begin{figure}[t!]
    \centering
    \includegraphics[width=2.3in]{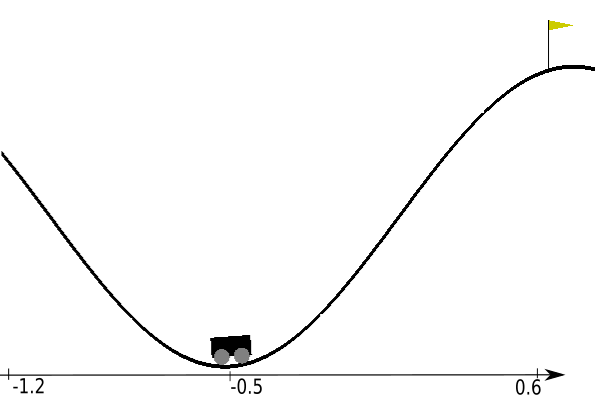}
    \caption{The mountain car environment. The shown position of the car at $-0.5$ is the starting position in our evaluations.}
    \label{fig:mountaincar}
\end{figure}

We experiment with the Mountain Car problem, where an agent needs to drive the car up the mountain in 1D by throttling left or right, as shown on Figure~\ref{fig:mountaincar}. The top of the mountain can only be reached by first building up momentum before throttling right. The agent spawns at a random position, and only observes a noisy position sensor and has no access to its current velocity. At each timestep, the agent can choose between two policies: $\pi_l$ to throttle to the left, $\pi_r$ to throttle to the right. We experiment with two flavors of this environment: one where the agent starts with fixed zero velocity, and one where the agent starts with a random initial velocity.

For our generative model, we instantiate $p_{\theta}(\vs_t | \vs_{t-1}, \va_{t})$, $p_{\phi}(\vs_t | \vs_{t-1}, \va_{t}, \vo_t)$ and $p_{\xi}(\vo_t | \vs_{t})$ as fully connected neural networks with 20 hidden neurons, and a state space with 4 dimensions. To bootstrap the model, we train on actions and observation of a random agent minimizing the loss function in Equation~(\ref{eq:model-obj}) using stochastic gradient descent. Next, we instantiate an active inference agent that uses Equation~(\ref{eq:Gest}) to plan ahead and select the best policy. As preferred state distribution $P(\vs_\tau)$, we manually drive the car up the mountain and evaluate the model's posterior state at the end of the sequence $\hat{\vs}_{end}$, and set $P(\vs_\tau)= \mathcal{N}(\hat{\vs}_{end}, 1)$. To limit the computations, the active inference agent plans ahead for 90 timesteps, allowing to switch policy every 30 timesteps, effectively evaluating a search tree with depth 3, using 100 samples for each policy ($K=30, D=3, N=100$). 

\begin{figure*}
    \centering
    \subfloat[left-right-right]{{\label{fig:randv_lrr}\includegraphics[width=1.5in]{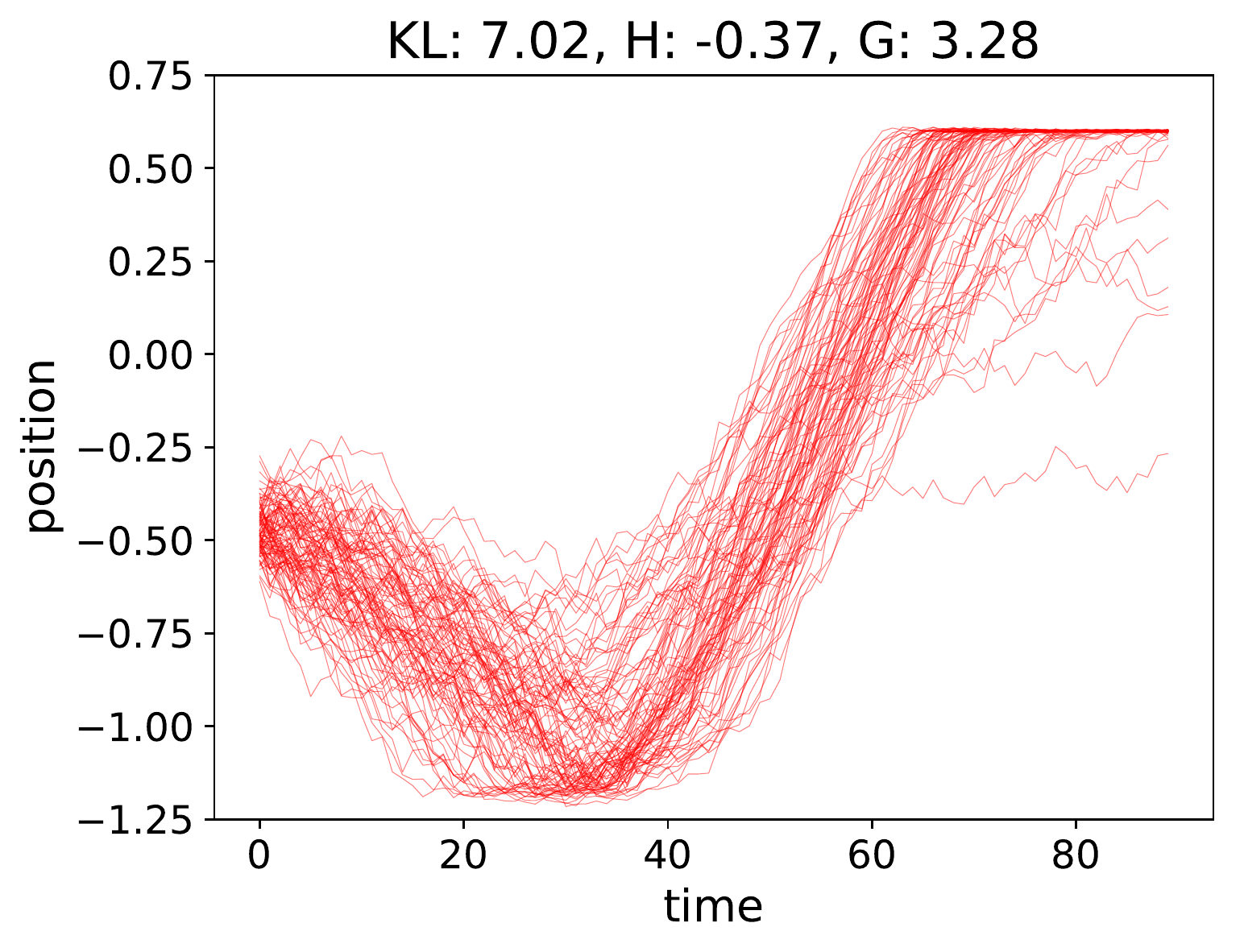}}}
    \subfloat[left-right-left]{{\label{fig:randv_lrl}\includegraphics[width=1.5in]{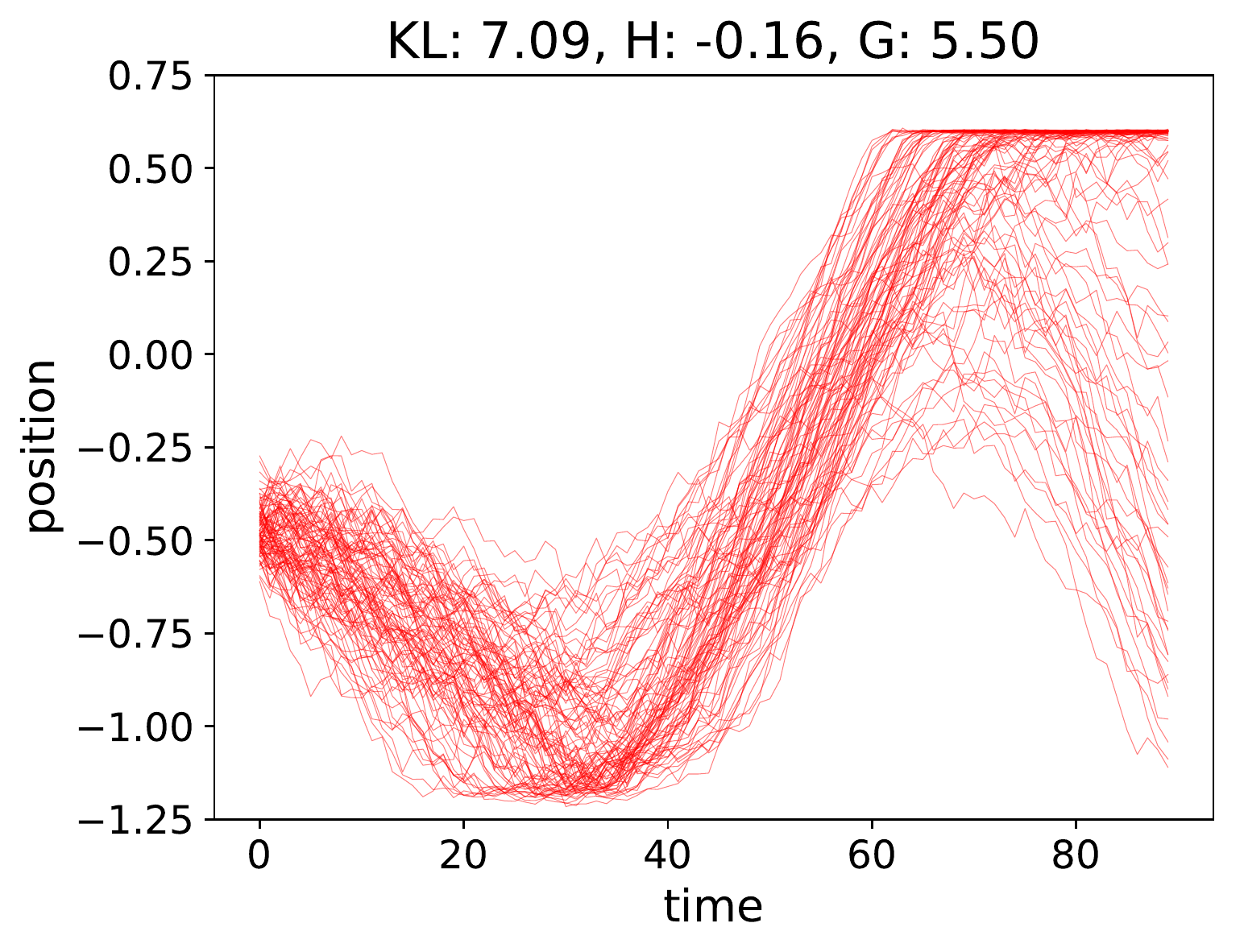}}}
    \subfloat[left-left-right]{{\label{fig:randv_llr}\includegraphics[width=1.5in]{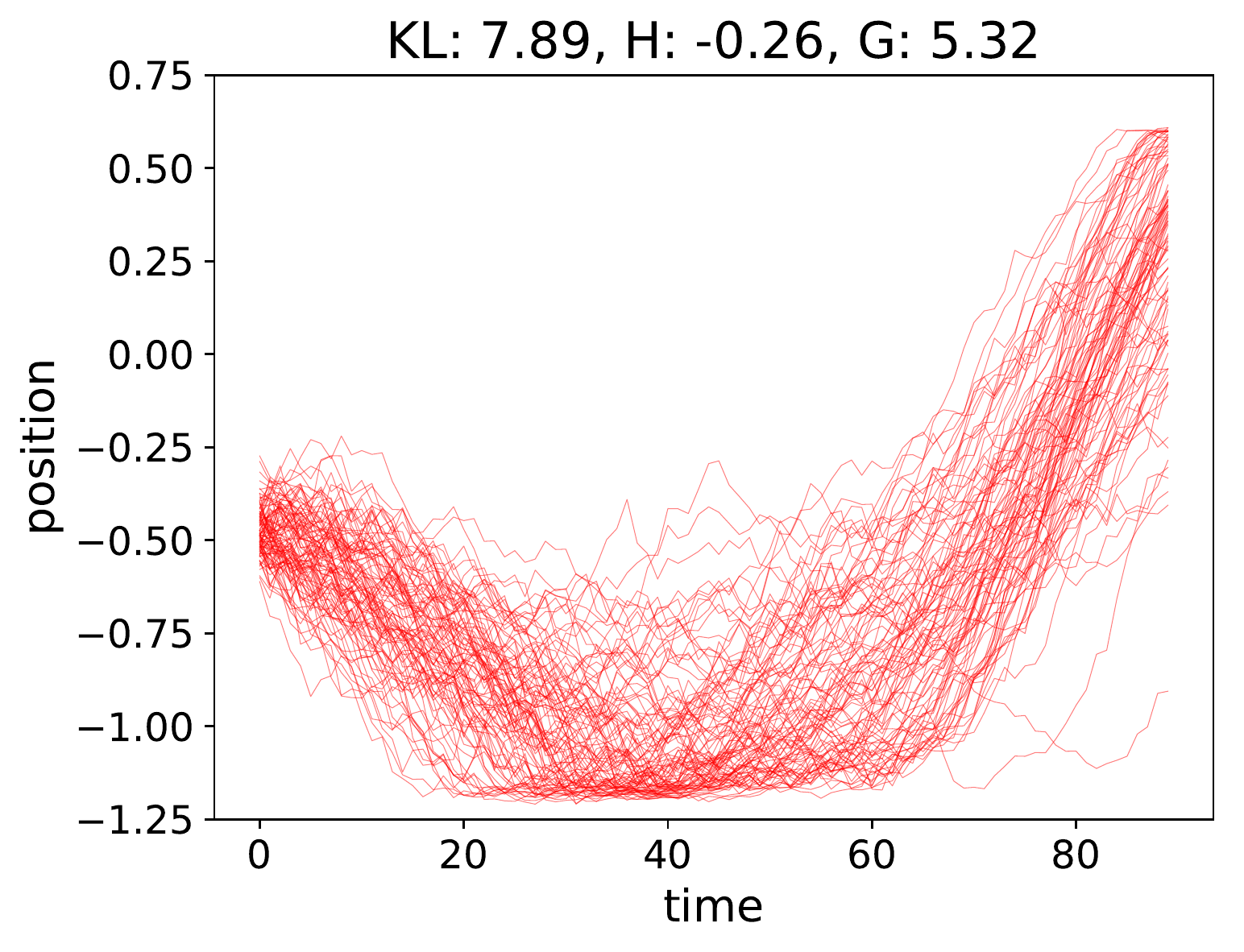}}}
    \subfloat[left-left-left]{{\label{fig:randv_lll}\includegraphics[width=1.5in]{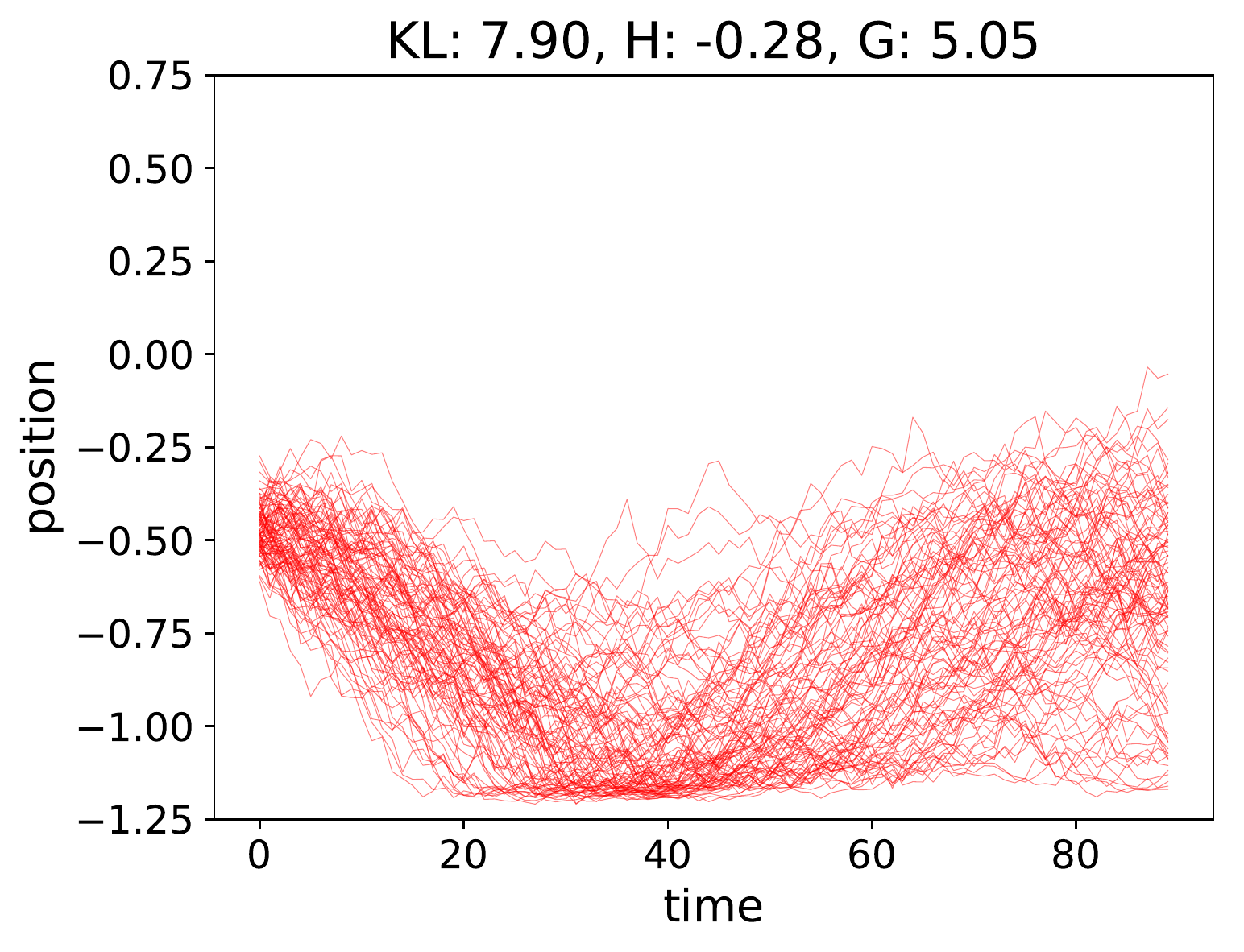}}}\\
    \subfloat[right-right-right]{{\label{fig:randv_rrr}\includegraphics[width=1.5in]{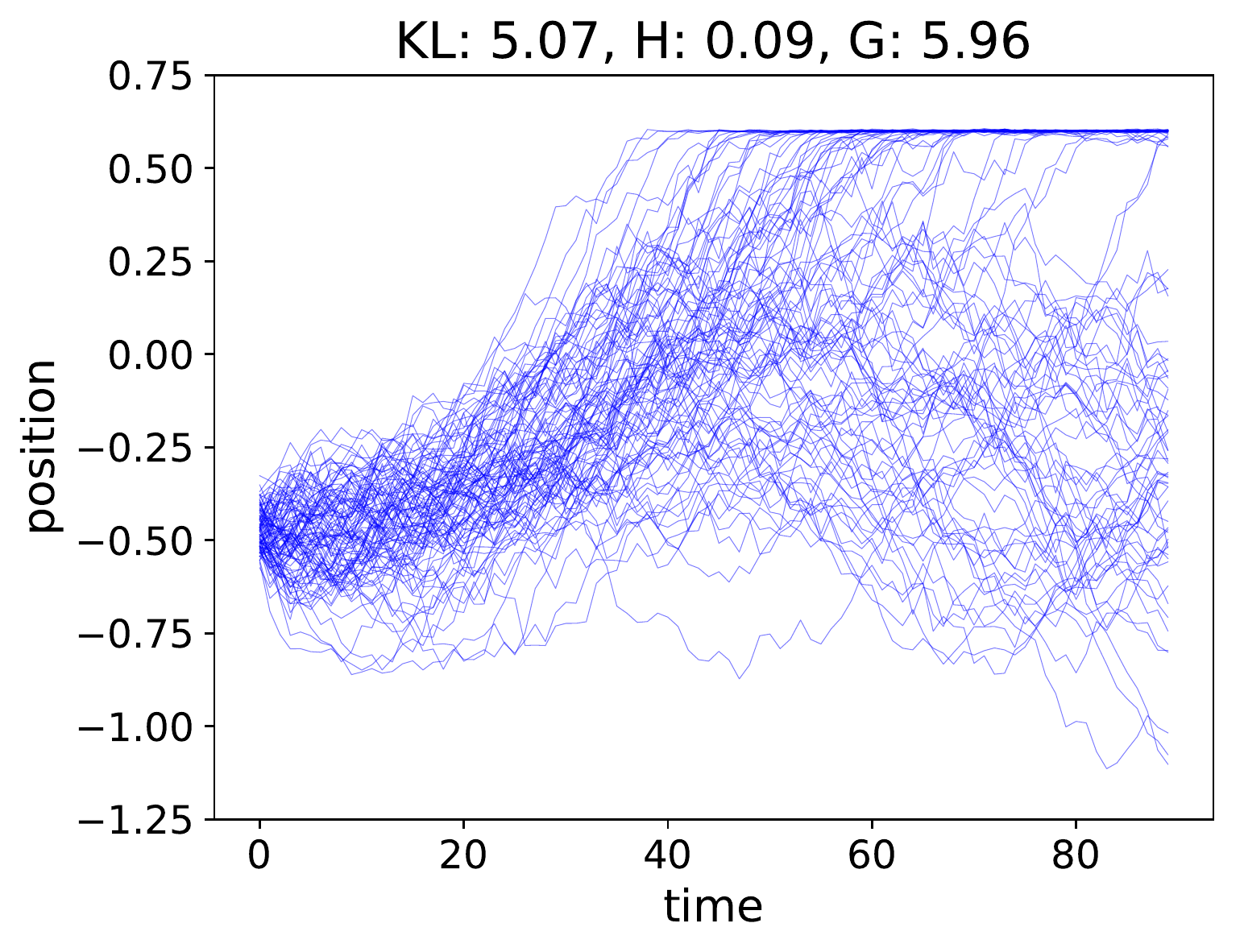}}}
    \subfloat[right-right-left]{{\label{fig:randv_rrl}\includegraphics[width=1.5in]{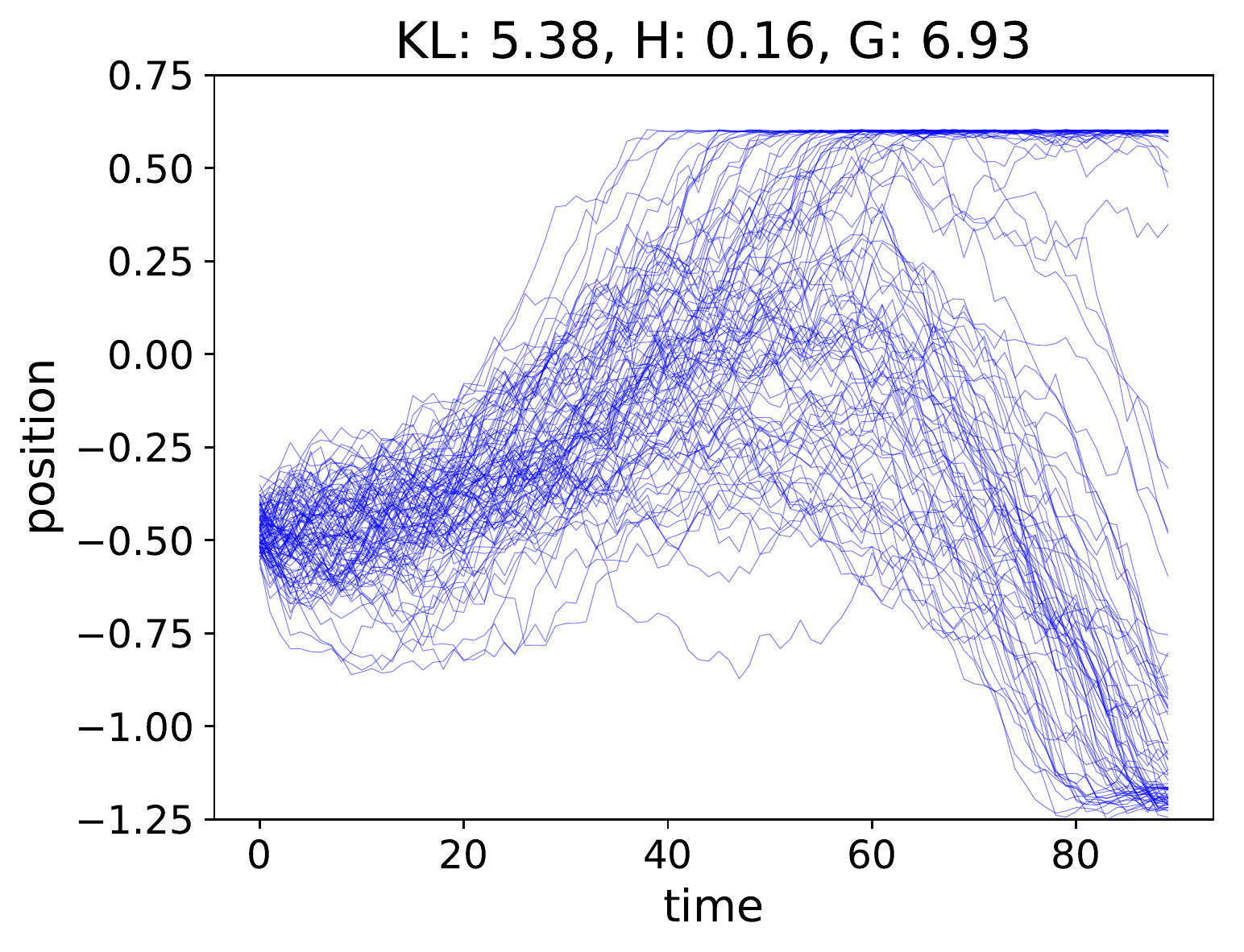}}}
    \subfloat[right-left-right]{{\label{fig:randv_rlr}\includegraphics[width=1.5in]{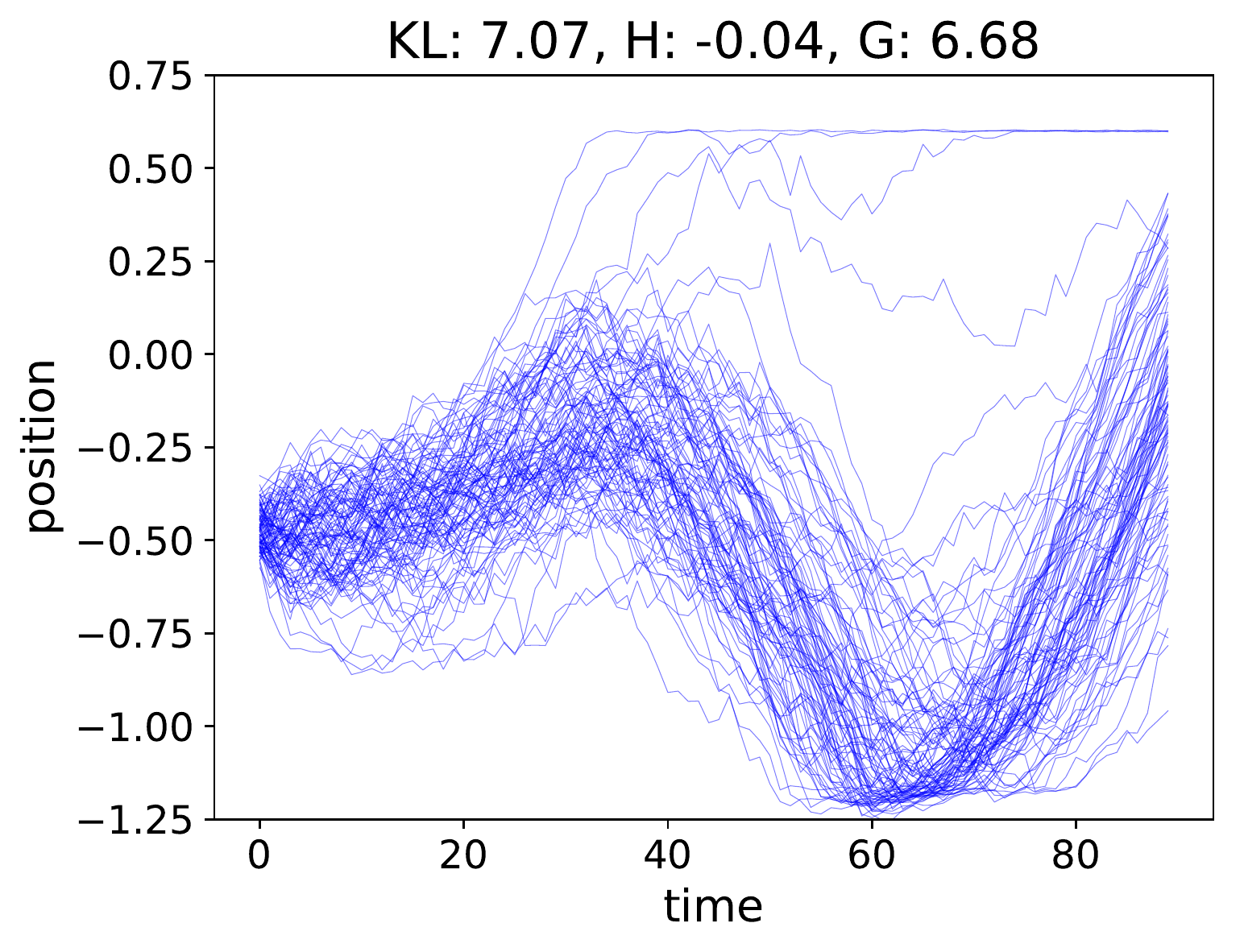}}}
    \subfloat[right-left-left]{{\label{fig:randv_rll}\includegraphics[width=1.5in]{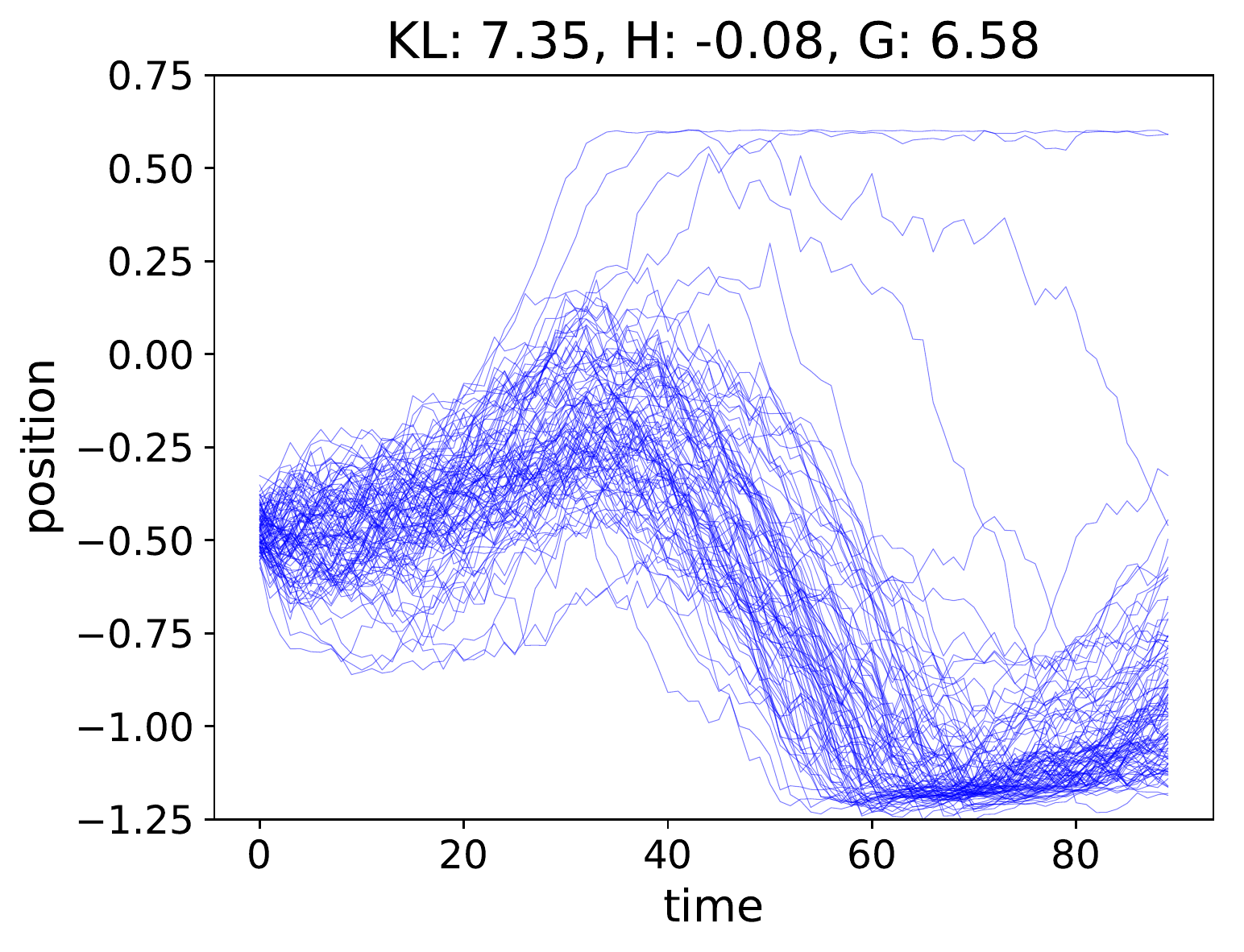}}}\\    
    \caption{Depending on the random initial velocity, the car will reach the hill fast using the right policy only part of the cases (e), however starting with the left policy first also reaches the hill top and with lower entropy on the trajectories (a). A greedy agent ($\rho > 1$) will pick (e) whereas a cautious agent ($\rho << 1$) will favor (a). For each policy we report the values of KL, H and G for $\rho = 0.1$.}
    \label{fig:randv}
\end{figure*}

Figure~\ref{fig:randv} shows the sampled trajectories for all branches of the search tree, in the case the model is bootstrapped with only a single observation at position $-0.5$. This is a challenging starting position as the car needs enough momentum in order to reach up the hill from there. In the case of a random starting velocity, the generative model is not sure about the velocity after only the first observation. This is reflected by the entropy (i.e. the expected ambiguity) of the sampled trajectories. Now following $\pi_r$ from the start will sometimes reach the preferred state, depending on the initial velocity. In this case the active inference agent's behavior is determined by the parameter $\rho$. For $\rho > 1$, the agent will act greedily, preferring the policy that has a chance of reaching the top early, cf.~Figure~\ref{fig:randv_rrr}. When setting $\rho << 1$, the entropy term will play a bigger role, and the agent will select the policy that is less uncertain about the outcomes, rendering a more cautious agent that prefers a more precise and careful policy, moving to the left first – see Fig.~\ref{fig:randv_lrr}. We found setting $\rho = 0.1$ yields a good trade-off for the mountain car agent.

\begin{figure*}
    \centering
    \subfloat[left-right-right]{{\label{fig:zerov_lrr}\includegraphics[width=1.5in]{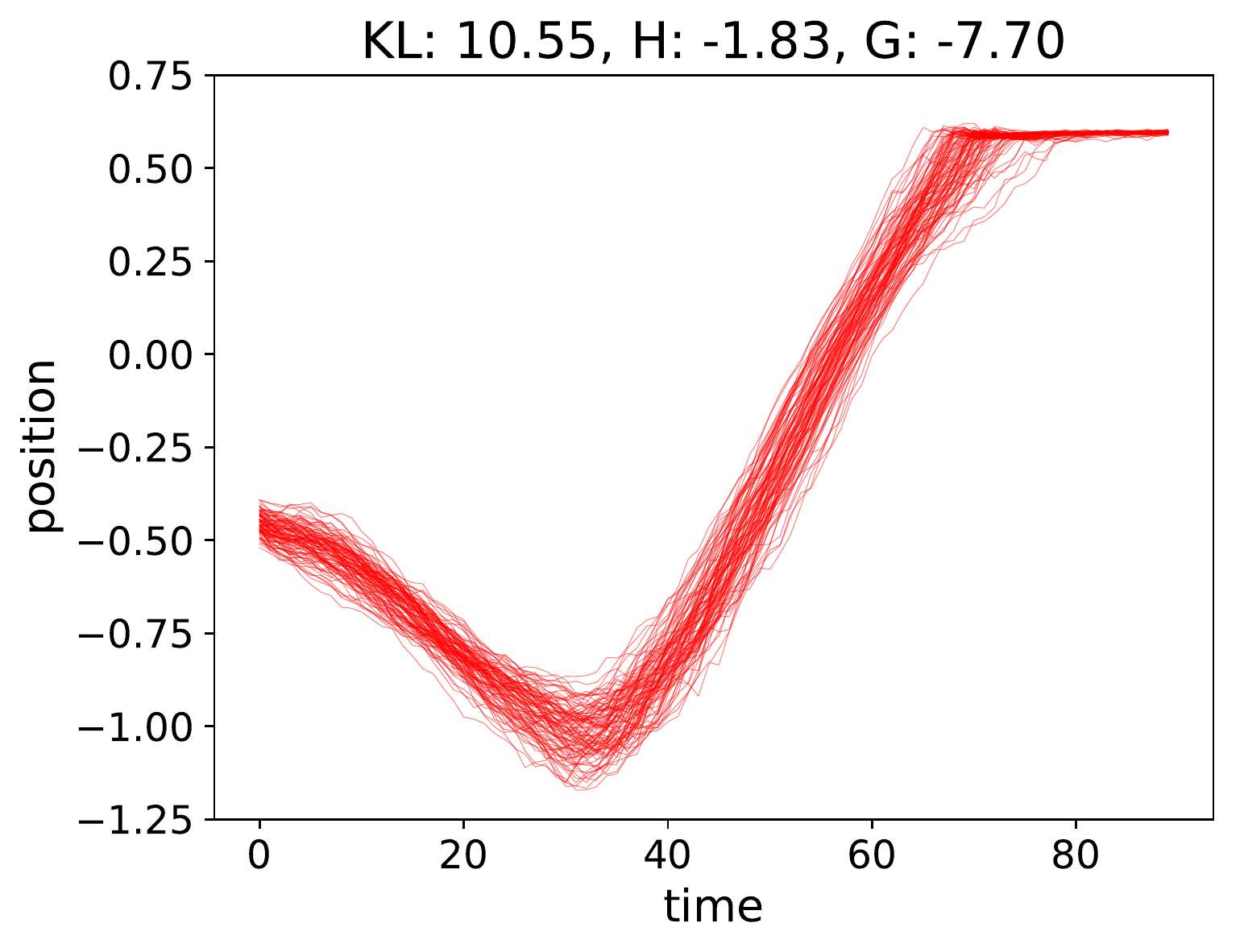}}}
    \subfloat[left-right-left]{{\label{fig:zerov_lrl}\includegraphics[width=1.5in]{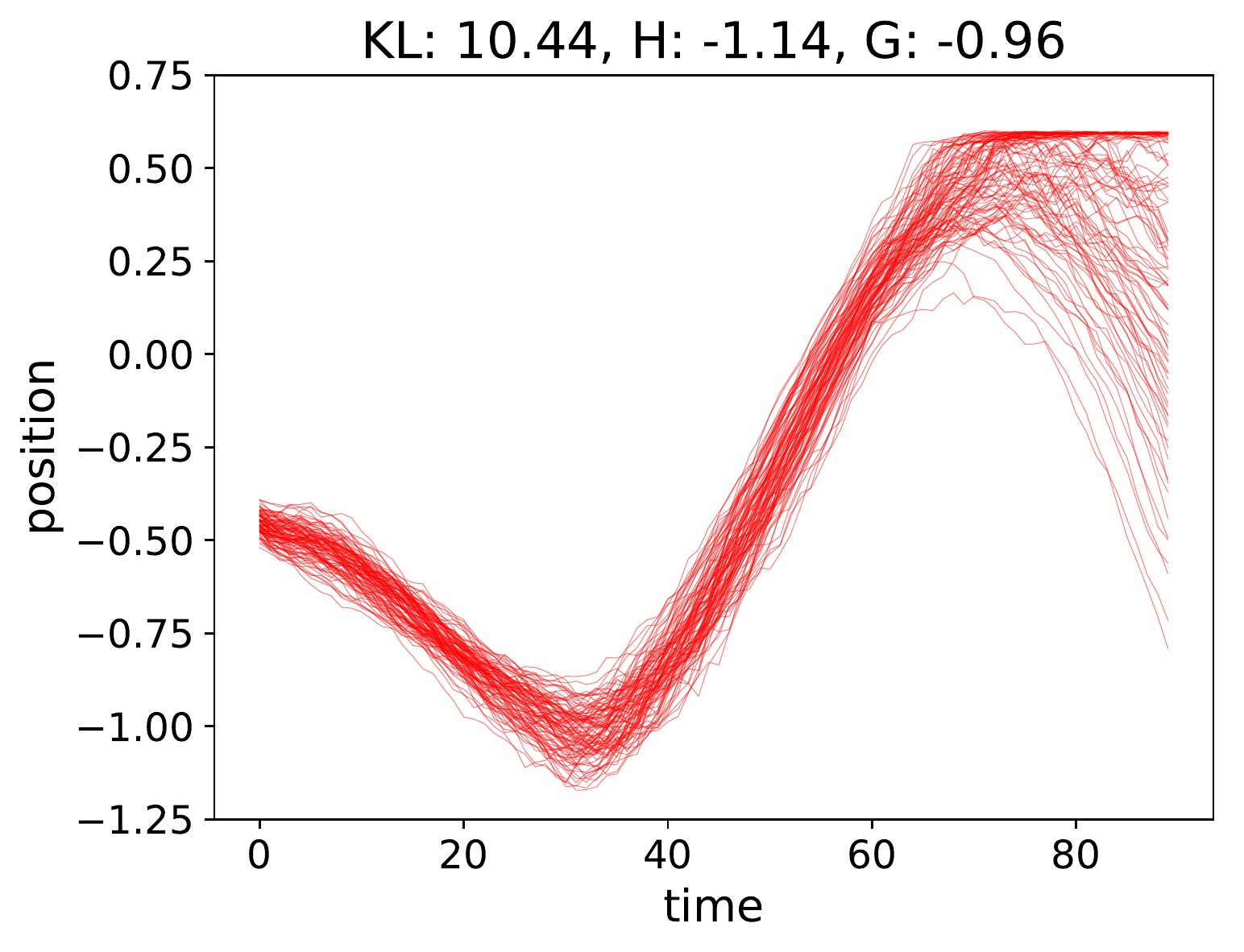}}}
    \subfloat[left-left-right]{{\label{fig:zerov_llr}\includegraphics[width=1.5in]{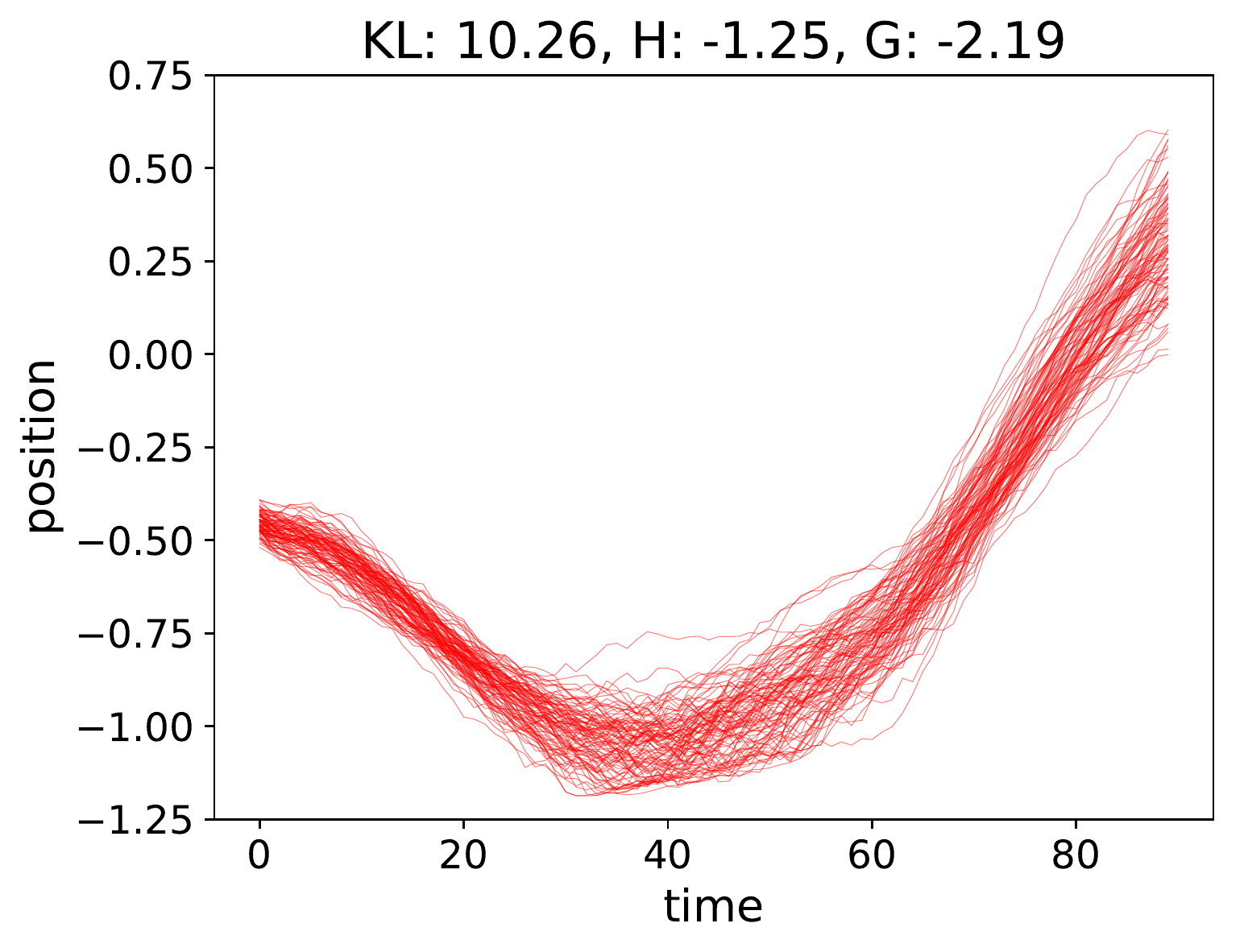}}}
    \subfloat[left-left-left]{{\label{fig:zerov_lll}\includegraphics[width=1.5in]{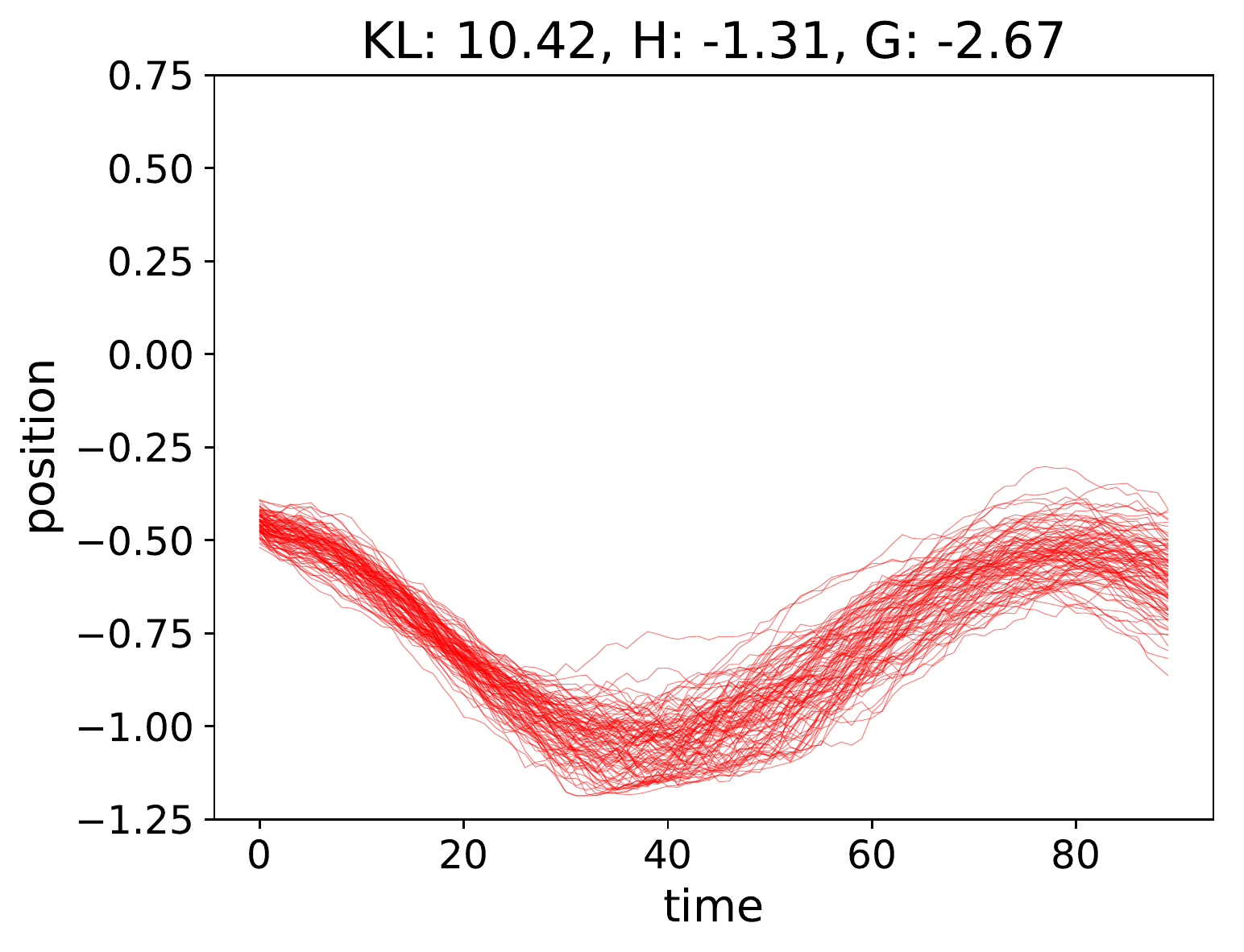}}}\\
    \subfloat[right-right-right]{{\label{fig:zerov_rrr}\includegraphics[width=1.5in]{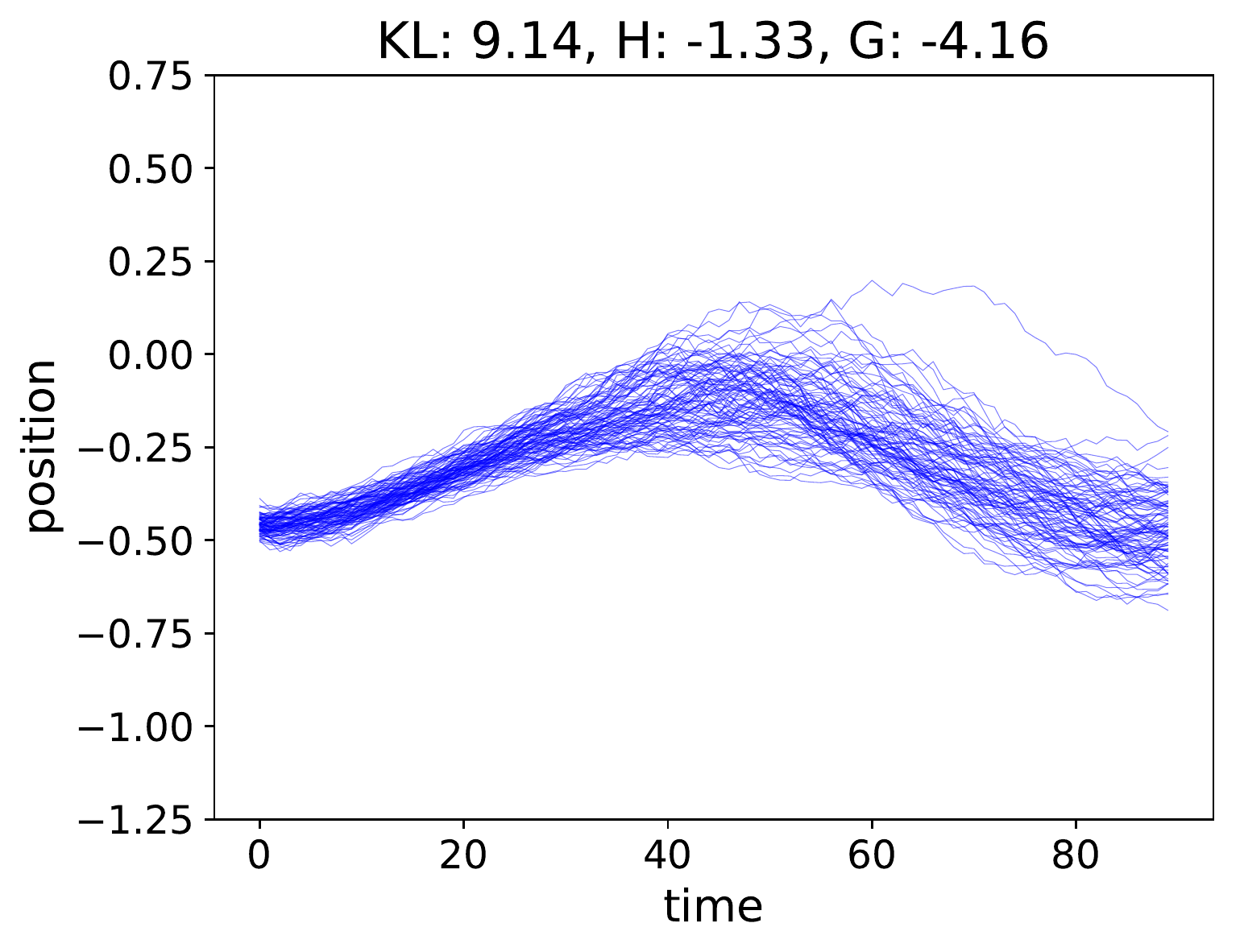}}}
    \subfloat[right-right-left]{{\label{fig:zerov_rrl}\includegraphics[width=1.5in]{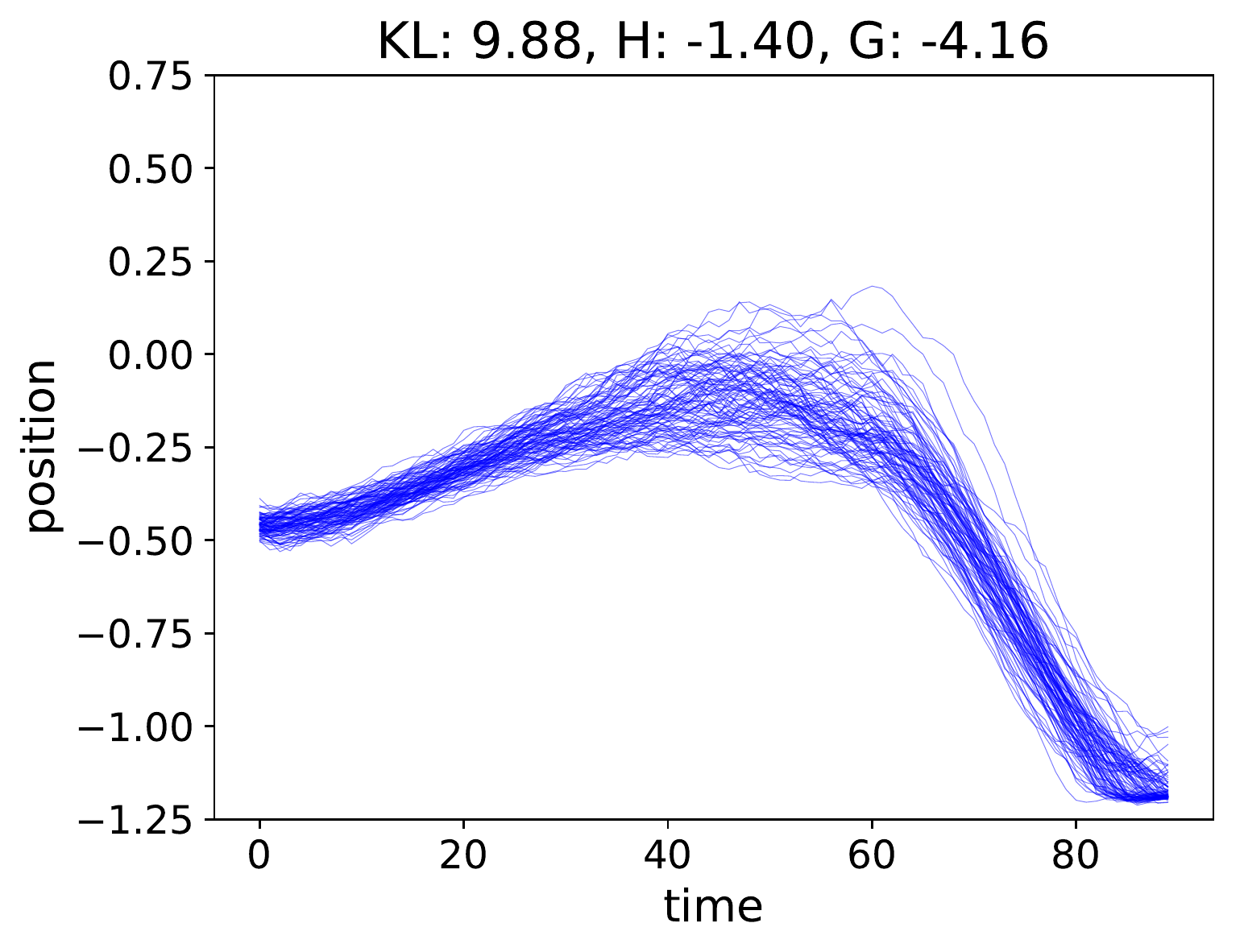}}}
    \subfloat[right-left-right]{{\label{fig:zerov_rlr}\includegraphics[width=1.5in]{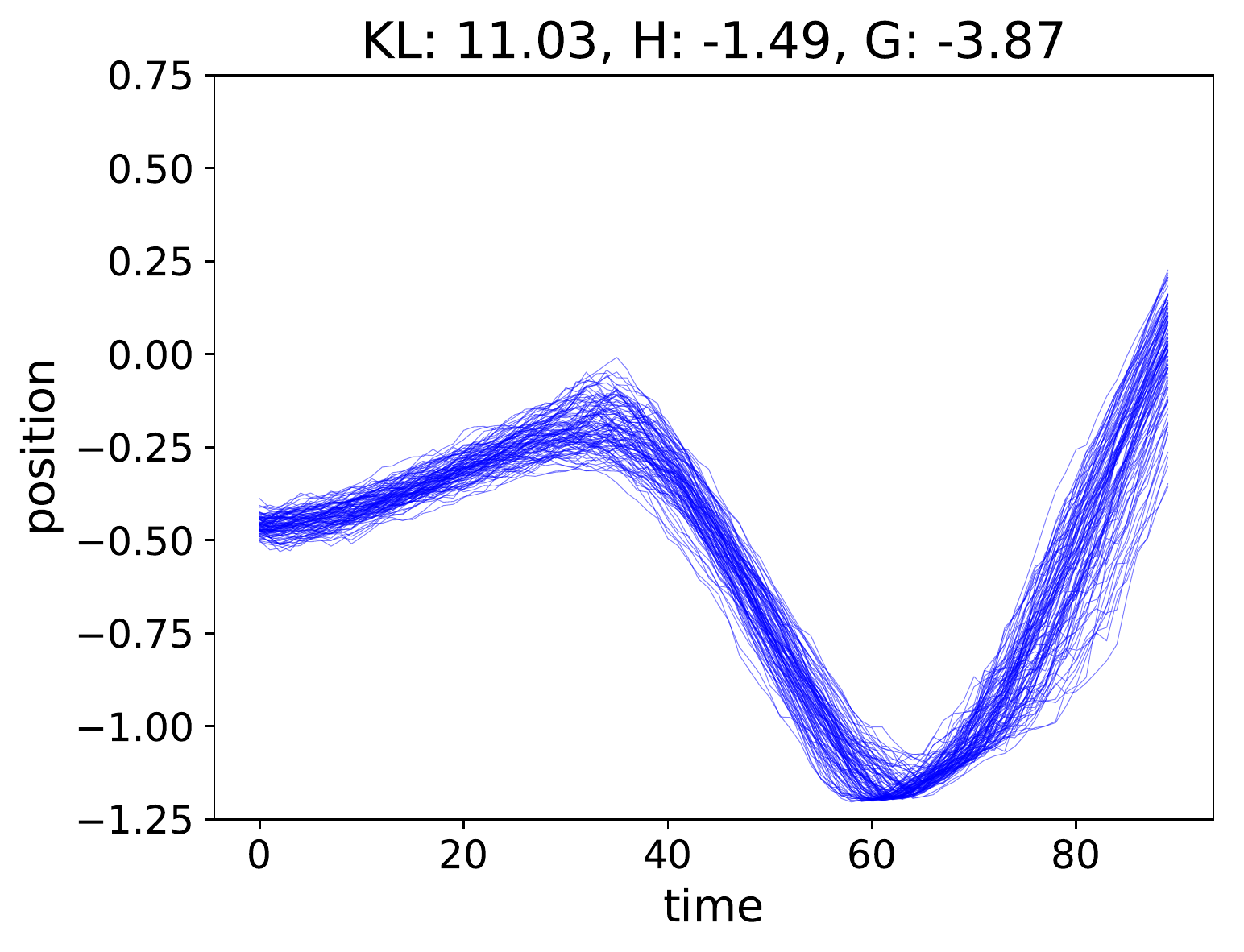}}}
    \subfloat[right-left-left]{{\label{fig:zerov_rll}\includegraphics[width=1.5in]{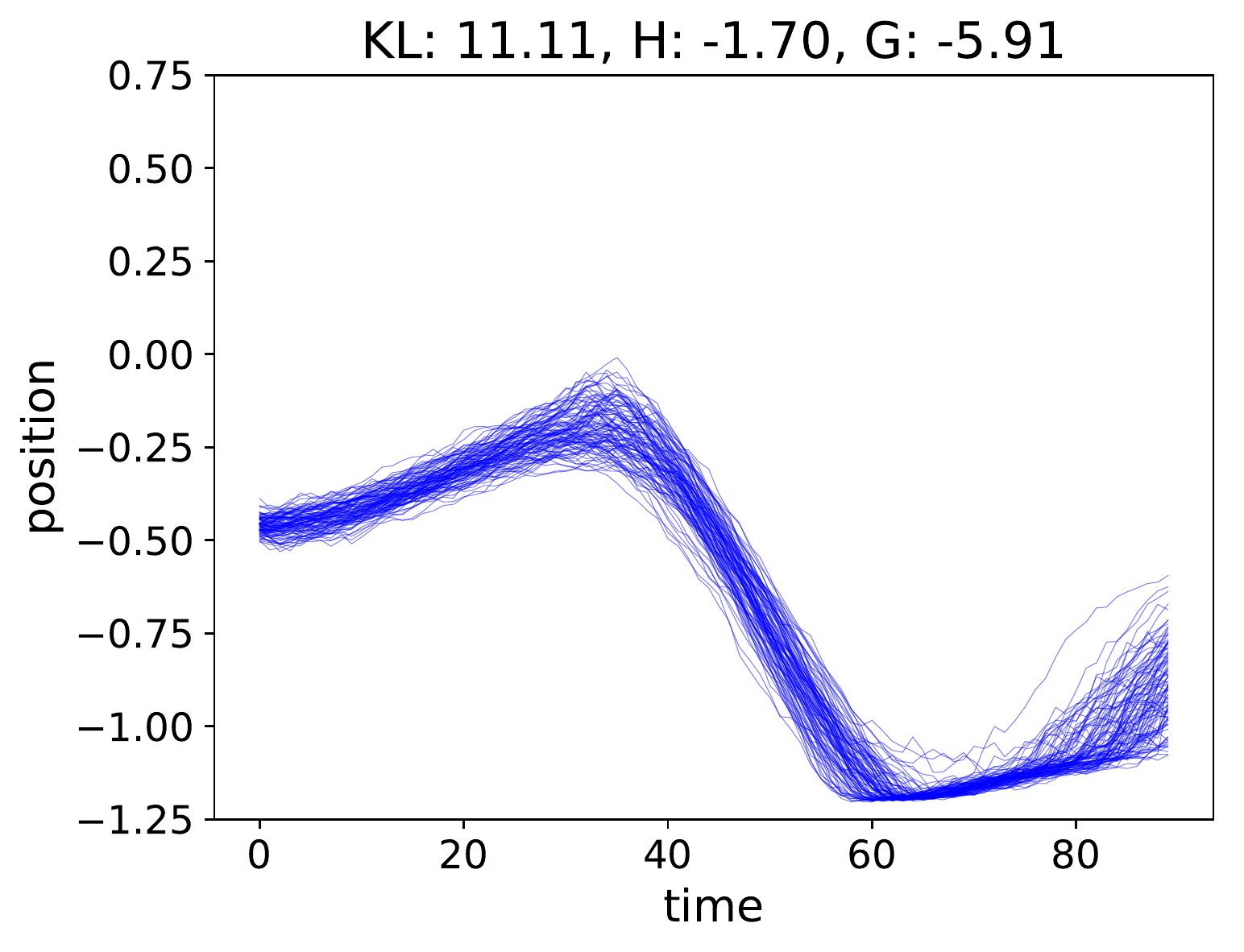}}}\\    
    \caption{When the environment starts the car with a fixed zero velocity, the model is on average much more certain on the predicted trajectories, resulting in lower entropy terms. However, policy (e) still achieves the lowest KL value, as this term is evaluated each timestep, and moving away from the preferred state yields a high KL penalty. When choosing $\rho = 0.1$, the agent again favors (a). For each policy we report the values of KL, H and G for $\rho = 0.1$.}
    \label{fig:zerov}
\end{figure*}

In the environment with no initial velocity, the transition model learnt by the agent is quite accurate and the entropy terms are an order of magnitude lower, as shown in Figure~\ref{fig:zerov}. However, in terms of preferred state the lowest KL is still achieved by following $\pi_r$. This is due to the fact that the KL term is evaluated each timestep, and moving to the left, away from the preferred state in the sequence outweighs the benefit of reaching the preferred state in the end. Choosing $\rho = 0.1$ again forces the agent to put more weight on resolving uncertainty, preferring the policy in Figure~\ref{fig:zerov_lrr}. 

\section{Discussion}
\label{sec:discussion}

Using deep neural networks to instantiate the generative model and to approximate both prior and posterior distributions, has the advantage that the generative model is independent of any state representation. The model can learn the best state representation for the observed data. Employing deep neural networks also opens up the possibility of using high-dimensional sensor inputs, e.g.~images. The downside of our model, however, is the required sampling step, which means that a distribution is only calculated for the next timestep, and distributions for timesteps $\tau$ further in the future can only be approximated by sampling. 

Another point of discussion is the definition of the preferred state distribution. In our case we opted for a Gaussian state distribution, centered around the state visited by an expert demonstration, similar to our earlier work~\cite{DeBoom2018}. However, the standard deviation of this distribution will determine the absolute value of the KL term in Equation~(\ref{eq:Gest}). A small standard deviation will blow up the KL term, completely ignoring the entropy term. A large standard deviation will assign probability mass to neighboring states, possibly introducing local optima that don't reach the actual goal state. To mitigate this, we introduced an additional $\rho$ parameter that balances risk and ambiguity.

Finally, planning by generating and evaluating trajectories of the complete search tree is computationally expensive. In this paper, we intentionally pursued this approach in order to directly investigate the effect of the KL term versus the entropy term. To mitigate the computational load, one might amortize the resulting policy by training a policy neural network $p_{\pi}(\va_t | \vs_t)$ based on the visited states and the planned actions by the agent, similar to~\cite{Catal2019}. In other approaches, the policy is learned directly through end-to-end training. For example, K.~Ueltzh{\"o}ffer uses evolution strategies to learn both a model and a policy, that requires part of the state space to be fixed, to contain information of the preferred state~\cite{Ueltzhöffer2018}. B.~Millidge on the other hand amortizes the expected free energy $G$ as function of the state, similar to value function estimation in reinforcement learning~\cite{Millidge2019}. Again, however, the perception part is omitted and the state space is fixed upfront.

\section{Conclusion}
\label{sec:conclusion}

In this paper, we have shown how generative models parameterized by neural networks are trained by minimizing the free energy, and how these can be exploited by an active inference agent to select the optimal policy. We will further extend our models to work in more complex environments, in particular towards more complex sensory inputs such as camera, lidar or radar data. 

%
%
%


\bibliographystyle{IEEEbib}
\bibliography{strings,refs}
\balance
\end{document}

%% file: math_commands.tex
\usepackage{amsmath,amsfonts,bm}

\def\eqref#1{equation~\ref{#1}}

\def\1{\bm{1}}

\def\va{{\bm{a}}}

\def\vo{{\bm{o}}}

\def\vs{{\bm{s}}}

\DeclareMathAlphabet{\mathsfit}{\encodingdefault}{\sfdefault}{m}{sl}
\SetMathAlphabet{\mathsfit}{bold}{\encodingdefault}{\sfdefault}{bx}{n}

\newcommand{\E}{\mathbb{E}}

\newcommand{\KL}{D_{\mathrm{KL}}}

